\documentclass[a4paper,fleqn,twocolumn,3p,preprint]{cas-dc}
\usepackage[authoryear,longnamesfirst]{natbib}
\usepackage{tabularx}
\usepackage{graphicx}
\usepackage{subcaption}
\usepackage[switch]{lineno}

\begin{document}
\begin{sloppypar}
\let\WriteBookmarks\relax
\def\floatpagepagefraction{1}
\def\textpagefraction{.001}

\title [mode = title]{Establish seedling quality classification standard for Chrysanthemum efficiently with help of deep clustering algorithm}

\author[1]{Yanzhi Jing}

\affiliation[1]{
    organization={College of Horticulture},
    addressline={South China Agricultural University}, 
    city={GuangZhou},
    postcode={}, 
    state={GuangDong},
    country={China}
}

\author[2]{Hongguang Zhao}

\affiliation[2]{
    organization={School of Computer Science and Technology, Faculty of Electronic and Information Engineering},
    addressline={Xi'an Jiaotong University}, 
    city={Xi'an},
    postcode={}, 
    state={Shaanxi},
    country={China}
}

\author[1]{Shujun Yu}
\cormark[1]
\ead{yushujun@scau.edu.cn}

\begin{abstract}
Establishing reasonable standards for edible chrysanthemum seedlings helps promote seedling development, thereby improving plant quality. However, current grading methods have the several issues. The limitation that only support a few indicators causes information loss, and indicators selected to evaluate seedling level have a narrow applicability. Meanwhile, some methods misuse mathematical formulas. Therefore, we propose a simple, efficient, and generic framework, SQCSEF, for establishing seedling quality classification standards with flexible clustering modules, applicable to most plant species. In this study, we introduce the state-of-the-art deep clustering algorithm CVCL, using factor analysis to divide indicators into several perspectives as inputs for the CVCL method, resulting in more reasonable clusters and ultimately a grading standard $S_{cvcl}$ for edible chrysanthemum seedlings. Through conducting extensive experiments, we validate the correctness and efficiency of the proposed SQCSEF framework.
\end{abstract}

\begin{keywords}
Seedling Quality Classification \sep K-Means Clustering \sep Deep Clustering \sep Chrysanthemum \sep Factor Analysis
\end{keywords}

\maketitle


\section{introduction}
Chrysanthemum is one of the most popular flower in the world(\cite{Spaargaren2018}). With the advancement of modern medical and chemical technology, researchers have found that edible chrysanthemum is rich in functional health ingredients(\cite{ZHENG2021127940}), such as a variety of vitamins, minerals and amino acids, chlorogenic acid, quercetin and baicalin, etc(\cite{molecules17066672}). The beneficial effects of Chrysanthemum are primarily attributed to its phenolic bioactive compounds, such as flavonoids and phenolic acids(\cite{tian2018protective}). These compounds are believed to possess antibacterial, antiviral, anti-inflammatory, and antioxidant properties, as well as free radical scavenging capabilities. They contribute to cardiovascular protection, prevention of coronary heart disease, cholesterol and lipid reduction, and benefit for diabetes management (\cite{yamamoto2015chrysanthemum}). Additionally, they have a positive impact on human health (\cite{hadizadeh2022chrysanthemum}).
Overall, edible Chrysanthemum is a multi-functional material that not only contains various nutrients but also promotes physical and mental health, alleviating some common health issues (\cite{lu2016phytochemical}).

Due to their various health benefits, edible Chrysanthemum has gradually become a highly favored health food (\cite{jiang2015flower}), available in a variety of forms (\cite{acikgoz2017edible}).
The growing demand for edible Chrysanthemum has made it one of the most common flowers in the edible flower market (\cite{doi:10.1080/87559129.2019.1639727}). This trend has spurred the development of related industries, with researchers focusing on improving Chrysanthemum quality, increasing yield, and exploring cultivation techniques suitable for different regions.

The establishment of seedling quality classification standards aims to ensure that the growth and yield of crops, horticultural plants, and forestry trees meet expected levels, thereby promoting the sustainable development of agriculture, horticulture, and forestry(\cite{sutton1980evaluation}).These standards not only ensure production quality and increase yield and quality but also enhance plant resistance to pests and diseases, promote varietal improvement, reduce production risks, regulate market order, and facilitate international trade(\cite{f10121064}).Overall, the implementation of seedling quality classification standards helps optimize production, protect the environment, and improve economic benefits, thereby laying a solid foundation for the sustainable development of agriculture and related industries.

Currently, numerous scholars have developed various indicators to assess seedling quality and performance potential, primarily based on morphological characteristics, e.g., plant height or ground diameter. These characteristics are easy to measure and widely applicable (\cite{pinto2011establishment}).However, traditional assessment methods primarily rely on manual inspection and basic indicators, which are highly subjective, labor-intensive, and inconsistent (\cite{zaerr1985role}).In practical production environments, seedling quality classification mainly relies on morphological indicators, while classification methods based on physiological indicators are still in the experimental research stage (\cite{grossnickle2018seedlings}).As seedling quality classification methods continue to evolve, it is increasingly necessary to further explore and refine a scoring system that comprehensively considers both morphological and physiological indicators (\cite{rose1990target}). Researchers need to establish a comprehensive and more applicable scoring system that considers both morphological and physiological indicators.

However, existing grading standard establishment methods have the following issues:
\begin{itemize}
    \item Only support a limited number of indicators. Clark et al. used visually selected indicators to establish grading standards for Northern Red Oak seedlings (\cite{10.1093/sjaf/24.2.93}). However,  describing plant development with only a few indicators inevitably results in information loss.
    \item The grading indicators are only applicable to specific species. 
    \item Some grading standard methods incorrectly use mathematical formulas. For example, principal component analysis (PCA) is designed for data dimensionality reduction, but a significant amount of work erroneously uses it for indicator selection.
\end{itemize}

To address these issues, we first defined a simple, efficient, and generic framework: Seedling Quality Classification Standard Establishment Framework, abbreviated as SQCSEF, which is suitable for establishing grading standards for most plants. Specifically, we first use clustering algorithms to divide the sample data into several clusters and compute the cluster centers and radii, then calculate the lower bounds of the clusters to determine the boundaries between grades. With the support of SQCSEF, we developed a grading standard for fresh chrysanthemum seedlings. Initially, we used the classic K-Means clustering method as the clustering module within SQCSEF, obtaining a grading standard $S_{kmeans}$ as a baseline. Subsequently, we introduced the state-of-the-art deep clustering algorithm CVCL, employing factor analysis to divide indicators into several views as inputs for the CVCL method, achieving more reasonable clustering results and ultimately a new grading standard $S_{cvcl}$. Upon comparison, we found that $S_{cvcl}$ is more precise and reasonable than $S_{kmeans}$, demonstrating that the SQCSEF framework is both efficient and generic. Moreover, the CVCL method effectively uncovers the intrinsic relationships between samples from different views. We summarize the contributions of this paper as follows:
\begin{itemize}
    \item To the best of our knowledge, SQCSEF is the first framework that support to establish  seedling quality classification for most variety of plants.
    \item We employ K-Means clustering method as the clustering module within SQCSEF, and obtain a grading standard $S_{kmeans}$ for edible
Chrysanthemum seedlings.
    \item We leverage CVCL, which is a deep clustering method, as the clustering module within SQCSEF. To meet the requirements of CVCL, we design a novel trick to partition samples into several views with help of factor analysis. Then we obtain a more precise standard $S_{cvcl}$.
\end{itemize}
The rest of the paper is organized as follows. In Section \ref{methods}, we first present materials for experiment and some basic mathematical preliminaries. The SQCSEF framework is designed in subsequent description. Then we introduce classic K-Means clustering and state-of-the-art deep clustering in detail. Section \ref{results} shows the experimental evaluation results. In the end, we conclude our work in Section \ref{conclusion}.

\section{Materials and methods}\label{methods}
\subsection{Experimental materials}
To ensure the consistency of plant growth, the materials utilized in this study were exclusively derived from Chrysanthemum morifolium cultivars, collected in March 2023 from the Zhongshan city. These were cultivated in the greenhouse at the College of Horticulture Experimental Base at South China Agricultural University, located at coordinates 113°21'30" E and 23°9'23" N. Propagation by cutting was initiated on March 10th, and data were collected on May 10th. During the cultivation period, the average annual temperature was observed to be between 19°C and 28°C, with precipitation levels ranging from 1,623.6mm to 1,899.8mm. The plant was identified as a member of the Chrysanthemum genus within the Asteraceae family.

We referred to the Technical Regulations for Cultivation of Fresh Edible Chrysanthemum Flowers in Zhongshan City (DB4420/T 18—2022) for plant cultivation and management methods. Furthermore, Liu Xiaobing et al. investigated the effects of different cultivation substrates on the growth of Huangqiu. Therefore, to ensure excellent water retention and air permeability of the cultivation substrate during the planting period of Huangqiu, we chose a mixture of peat soil, perlite, vermiculite, and pond mud in a ratio of 3:1:0.5:1.

To ensure good rooting effects and significant reduction in production costs, cultivation containers should possess characteristics such as environmental friendliness, low cost, and ease of use. Non-woven planting bags meet these criteria, so we selected them as the planting containers. During actual planting, Huangqiu seedlings were planted in non-woven planting bags, with bag dimensions of 30×25 cm and substrate filling up to 3/5 of the bag height. This ensures both good rooting effects and significant reduction in production costs. To prevent weed growth, maintain substrate air permeability, and avoid over-wetting of the planting bags, they were placed on a 1.5×3m iron rack, isolating the roots from the soil. 

Considering the preference of chrysanthemums for moist but not waterlogged conditions, an automatic sprinkler system was installed in the greenhouse. Due to the high temperatures from July to September, water evaporates quickly, so automatic watering was scheduled for 5 minutes every afternoon at 6 pm. After October, as temperatures decrease, the frequency of automatic watering was reduced to once every two days, with each session lasting for 8 minutes. In September, when nighttime temperatures are higher, there is a tendency for willow leaf buds to appear. If found, they should be promptly removed from above the normal leaves. To improve the quality of chrysanthemum flowers, solar supplementary lighting was installed above the greenhouse for one and a half months, from early October to mid-November, from 6 pm to 3 am.

Prior to the grading of seedlings, it is imperative to eliminate any plants afflicted with diseases or exhibiting poor growth. For comprehensive assessment, 200 species were randomly selected for measurement from each category, encompassing six parameters: seedling height, stem diameter,number of lateral branches, root length,  fresh weight, and leaf chlorophyll content. The specific measurement procedures were as follows: seedling height was measured using a measuring tape with a precision of 0.1 cm; stem diameter was assessed with an electronic digital caliper accurate to 0.01 mm; root length was likewise measured using a digital caliper, with precision to 0.01 mm; fresh weight was determined using an electronic scale, precise to 0.01 g; leaf chlorophyll content was quantified using the SPAD-502 chlorophyll meter. These measurement techniques were selected to ensure the accuracy and reliability of the data for subsequent analysis and evaluation.

\subsection{Basic preprocessing and statistical analysis of dataset}\label{preprocessing}
It is deserved to note that we ought to transform all metrics into maximization objectives before the preprocessing. For example, the seedling height and ground diameter are in category of maximization objectives since that they have a positive effect on seedling growth, while number of lateral branches is a minimization objective, which will consume nutrients of seedlings, block the sunlight, and increase the risk of diseases and pests. A simple formula to finish this work is shown as Eqs.\ref{eqs:forward}.
\begin{equation}\label{eqs:forward}
    x_i'=max(x)-x_i
\end{equation}
To compare and analyze indices with different scales fairly, the min-max normalization method is called at the beginning of preprocessing. Suppose that raw dataset $D_{raw}$ is a $M\times N$ matrix where each row is an element and each column corresponds to an index, we scale it to the standardized dataset $D_{std}$ according to Eqs.\ref{normalization}
\begin{equation}\label{normalization}
    e_{ij}^{'}=\frac{e_{ij}-min(e_j)}{max(e_j)-min(e_j)}
\end{equation}
where $e_{ij}$(resp., $e_{ij}^{'}$) is the value at the $i$-th row and $j$-th column of $D_{raw}$(resp., $D_{std}$).

Ideally, there is always some correlation between indices of plants, so we firstly attempt to introduce Pearson correlation coefficient(PCC) to describe correlation among indices. Assume that there are two variables $X$ and $Y$, the correlation coefficient $\rho$ is defined as
\begin{equation}
    \rho_{X,Y}=\frac{cov(X,Y)}{\sigma_X \sigma_Y}   
\end{equation}
where $cov(X,Y)$ is the covariance of $X$ and $Y$ which could be calculated according to Eqs.\ref{cov}, and $\sigma$ is the standard deviation.
\begin{equation}\label{cov}
    cov(X,Y)=\frac{ \underset{i=1}{\overset{M}{\sum}} \big(X_i-E(X)\big)\big(Y_i-E(Y)\big) } {M}
\end{equation}
Then we can obtain a correlation coefficient matrix. To estimate the robust of correlation coefficient, hypothesis testing should be conducted(\cite{2017i}). For simplicity, we show the calculation method of hypothesises testing directly. Define two hypothesis as
\begin{equation}\label{H0H1}
\begin{aligned}
    & H_0: r=0\\
    & H_1: r\neq 0
\end{aligned}
\end{equation}
where $r$ is the correlation coefficient between two variables $X$ and $Y$. Construct a new variable $t$ as
\begin{equation}
    t=r\sqrt{\frac{M-2}{1-r^2}}
\end{equation}
which actually obeys the t-distribution with $M-2$ degree of freedom(DOF). Figure out the probability $p(t)$ at value $t$ in t-distribution mentioned before, then the significance level of correlation coefficient is revealed. For example, if we calculate $p$ of some variable less than 0.05, the hypothesis $H_0$ is rejected at 95\% confidence level, in other word, the correlation coefficient is not equal to $0$ at 95\% significance level.

Note that the dataset ought to meet two requirements if use PCC: (1) the correlation type between two variables must be linear correlation and (2) samples of each variable should obey a normal distribution. Hence we need to do some inspection before calculating PCC. To briefly examine the correlation type, we choose to plot a scatterplot matrix which represent correlation between any two variables vividly. It is slightly complicated to check the distribution type that a variable obey. Jarque-Bera testing is conducted to test if a variable obeys normal distribution. Define two hypothesises as
\begin{equation}
    \begin{aligned}
        & H_0:X\ \rm{obeys\ normal\ distribution} \\
        & H_1:X\ \rm{doesn't\ obey\ normal\ distribution}
    \end{aligned}
\end{equation}
and construct a new variable $J\!B$ as
\begin{equation}
    J\!B=\frac{n}{6} \big[sk^2+\frac{(ku-3)^2}{4}\big]
\end{equation}
where $sk$ and $ku$ is the skewness and kurtosis of $X$ respectively. If $X$ obeys normal distribution, variable $J\!B$ must obey the chi-square distribution with $2$ DOF, i.e., $J\!B \sim \mathcal{X}^2(2)$. For example, we can figure out the probability $p(J\!B)$ in chi-square distribution mentioned before, and test if $p(J\!B)$ is larger than $0.01$. The $p(J\!B)>=0.01$ indicates that $X$ obeys normal distribution at 99\% confidence level.·

When the correlation between two variables is non-linear, PCC should be replaced by Spearman correlation  coefficient(SCC) which is defined as
\begin{equation}
    r_s=1-6\frac{\underset{i=1}{\overset{M}{\sum}} d_i^2}{M(M^2-1)}
\end{equation}
where $d_i$ is the rank difference between sample $X_i$ and $Y_i$. Similar to the robust estimation phase of PCC, we still use hypothesises the same as Eqs.\ref{H0H1} shown, and construct a new variable $st$ according to Eqs.\ref{st}.
\begin{equation}\label{st}
    st=r_s\sqrt{M-1}\sim N(0,1)
\end{equation}
Then we figure out the probability $p(st)$ in standard normal distribution and compare it with $0.05$. If $p(st)\textless 0.05$, the correlation coefficient is not equal to 0 at 95\% significance level.

\subsection{Framework of establishing seedling quality classification standard}\label{framework}
In recent years, some studies on seedling quality classification standard for forests, cash crops, flowers and herbs have been proposed(cite papers). However, there doesn't exist a framework that could be employed to establish the standard for any plant. So we attempt to define a simple, efficient and universal framework(namely SQCSEF, i.e., seedling quality classification standard establishment framework) to address this issue. Assuming that we have done the preprocessing work of raw dataset $D_{raw}$ as we described in section \ref{preprocessing}, and obtained the dataset $D_{std}$ containing $M$ elements. The element is a $N$-dimensional vector where each dimension actually is an attribute. Firstly, SQCSEF takes $D_{std}$ as input and calls the clustering method to partition the dataset $D_{std}$ into $K$ clusters $C_1,C_2,...,C_K$ and figure out the centroid $c_i,i=1,2,...,K$ of each cluster. Note that the specific clustering method is chosen by the user, which means our framework is suitable for different varieties as long as calling appropriate clustering method. In practice, $K$ is likely to set to 3 corresponding to 3 levels of seedling quality. Then we calculate the radius of each cluster according to Eqs.\ref{radius}
\begin{equation}\label{radius}
    r_i=\sqrt{\Sigma_{k=1}^N \sigma_{ik}^2}
\end{equation}
where $\sigma_{ik}$ is the standard deviation of cluster $C_i$'s k-th attribute according to Eqs.\ref{standard deviation}. We use notion $M_i$ to denote the number of elements in cluster $C_i$, and notion $e[k]$ to denote the k-th attribute value of element $e$.
\begin{equation}\label{standard deviation}
    \sigma_{ik}=\sqrt{\frac{1}{M_i-1} \Sigma_{j=1}^{M_i} (e_j[k]-\overline{e[k]})^2 }
\end{equation}
After that, we calculate the lower bound of each level using radiuses and centroids of clusters. The traditional calculation method of lower bound draws a circle and straight line through point centroids on a chequered paper, then measures the point of intersection as lower bound. However, it has two limitations: poor precision and unable to support high dimensional data. To address this limitations, we derive a formula as shown in Eqs.\ref{lower bound} to support data of any dimension while maintaining high precision. The notion $lb_{ik}$ denotes the lower bound of cluster $C_i$'s $k$-th attribute.
\begin{equation}\label{lower bound}
    lb_{ik}=c_i[k]-\frac{c_i[k] \cdot r_i}{||c_i||_2}
\end{equation}

The lower bound is set as the boundary point between levels in classification standard. Note that $lb$ values are corresponding to the standardized dataset $D_{std}$, so we need to transform them to real value corresponding to raw dataset $D_{raw}$ according to Eqs.\ref{normalization}. In practice, we intend to follow the principle of half meeting, i.e., if half or more of the indices of a sample meet the level $X$, we will consider this sample belongs to level $X$. The remaining challenge is how to choose an appropriate and efficient clustering method. Aiming to compare the performance of traditional clustering and emerging deep clustering, we introduce two clustering methods, K-Means and CVCL in section \ref{traditional clustering} and section \ref{deepclustering} respectively.

\subsection{Traditional cluster analysis algorithm}\label{traditional clustering}
Cluster analysis(\cite{clustering}), which is also named as clustering, is an algorithm in category of unsupervised learning. Given a set of elements, cluster analysis is supposed to form some clusters such that for arbitrary cluster $C_i$ and element $e_j$ in $C_i$, $e_j$ is more similar to other elements in cluster $C_i$ than elements in other clusters. There are several distances to represent the similarity between elements, such as Euclidean distance and Hamming distance. During the processing, the algorithm tries to reduce the sum of distance between each elements in the cluster as less as possible until the algorithm converged. Cluster analysis is extensively utilized in knowledge discovery and data mining to partition the dataset into several clusters corresponding to some classes. For example, given  descriptions of apple, banana, cat and dog, the cluster analysis algorithm could classify apple and banana as cluster "Fruit" while classify cat and dog as cluster "Animal".

K-Means clustering is a wildly used cluster analysis algorithm proposed by Stuart Lloyd(\cite{1056489}). Recall that we have a dataset of $M$ elements $e_1,e_2,...,e_M$  and each element is a $N-$dimensional vector corresponding to $N$ attributes, the key idea of K-Means clustering is to partition these elements into $K$ clusters s.t.
\begin{equation}
    \mathop{\arg \min}\limits_{K} \Sigma_{i=1}^{K}\Sigma_{e\in C_i} || e-\hat{e} ||_2^2
\end{equation}
where $\hat{e}$ is the mean of elements in cluster $C_i$ and $||v||_2$ is the Euclidean norm of vector $v$. Evidently it is NP-hard to solve this multi-objective optimization problem, but K-Means clustering gets the local optimum intelligently using the Heuristic algorithm. More concretely, the algorithm choose $K$ elements from the dataset randomly as the initial $K$ centroids(Step I). Through assigning every elements to nearest cluster, we can obtain $K$ clusters of preliminary version(Step II). Then the algorithm updates $K$ centroids with mean of elements in each cluster(Step III). Step II and Step III are executed repeatedly until the difference between result of two successive iterations is below the Threshold value predefined by the user, and actually we get the local optimum.


\begin{figure*}[htbp]
	\centering
		\includegraphics[width=1\linewidth]{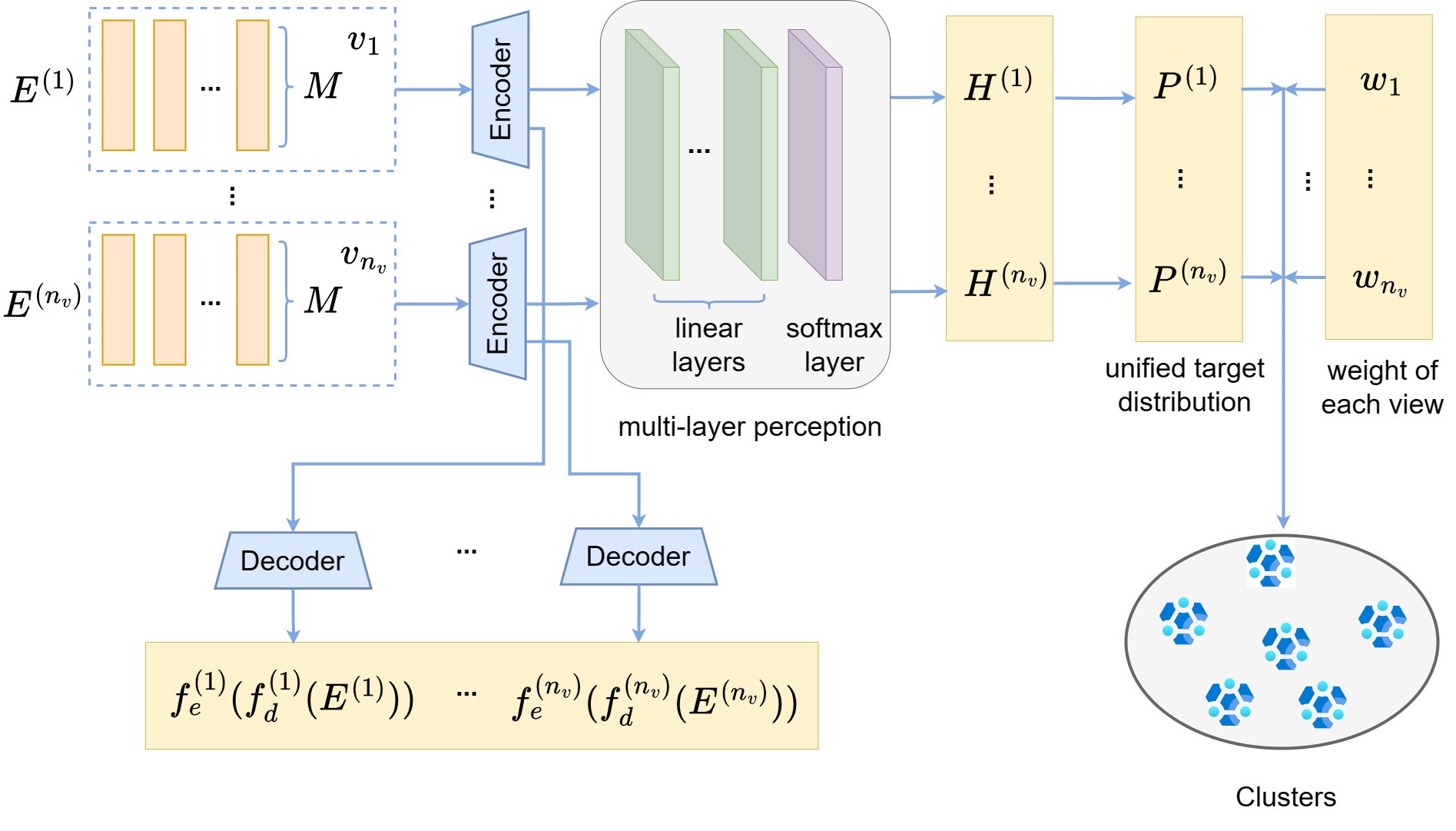}
	  \caption{CVCL method architecture. Here we introduce weight to each view.}\label{model}
\end{figure*}

\subsection{Deep Clustering}\label{deepclustering}
While K-Means clustering is efficient and easy to implement, it has one drawback that can not be ignored: it uses $||e-\hat{e}||_2^2$, i.e., Euclidean distance, as a measure of similarity, which could induce severe fault when some attributes is abnormal. For instance, supposed that we have an element $e_1=[2,2,10]$ and two centroids $c_1=[1,1,1],c_2=[8,8,8]$ of cluster $C_1,C_2$ respectively. Each dimension of vector represents seedling's Height, Ground diameter and Weight respectively. So we can calculate the Euclidean distance from $e_1$ to $c_1$ and $c_2$, i.e., $d_1=||e_1-c_1||_2^2=83$, $d_2=||e_2-c_2||_2^2=76$. Based on the algorithm of K-Means described above, $e_1$ should be assigned to cluster $C_2$ as $d_2\leq d_1$. However, the result is contrast to common sense since that when a seedling's Height and Ground diameter are lower than normal, it has a poor grouth even though weight of which has a good level. In fact, $e_1$ should be assigned to cluster $C_1$ in practice because of its poor Comprehensive grouth. 

Reviewing the role of K-Means clustering in the establishment of seedling quality classification standard, we found that it's mainly used to form three clusters corresponding to three levels of seedlings. Due to the potential negative impact of some abnormal attribute values in the sample dataset, K-Means may give unreasonable answers. So it is of vital significance to find an appropriate clustering algorithm to achieve higher accuracy. 

With the rapid development of Deep Learning, deep clustering is proposed to replace the traditional clustering methods which suffered from negative impact of abnormal elements and huge computational overhead on large-scale sample datasets. Xie et al. proposed Deep Embedded Clustering for the first time in 2016, the key idea of which is to map elements to lower-dimensional feature space using neural network(\cite{10.5555/3045390.3045442}). Then a sequence of methods were proposed, such as DEeP Embedded Regularized ClusTering proposed by Kamran et al. based on convolutional autoencoder and multinomial logistic regression function(\cite{8237874}), deep manifold clustering proposed by Chen et al. which utilized a clustering-oriented objective and a locality preserving objective as model's optimization objective(\cite{Chen2017UnsupervisedMC}), USNID proposed by Zhang et al. which designed a centroid-guided clustering mechanism(\cite{10349963}), and Secu proposed by Qi which adds a pre-training stage for representation learning(\cite{Qian_2023_ICCV}), etc.


Chen et al. proposed cross-view contrastive learning (CVCL) method in 2023 that concentrated on semantic label consistency of multi-view data(\cite{Chen_2023_ICCV}). The core idea of CVCL is  utilizing contrastive learning to extract common semantic labels from multi views. Obviously, CVCL method can address our issue suitably just with a few modifications since that we can divide indices of seedling into several views from different aspects: physiology view and morphology view, or aboveground part view and underground part view. For example, the leaf chlorophyll content belongs to physiology view while the height and ground diameter of seedlings belong to morphology view. To appropriately decide the views, the descending dimension method such as principal component analysis(PCA) or factor analysis(FA) is called before conducting CVCL method. Compared with PCA, factor analysis can identify correlations between latent factors and observed variables, and analysis result is easier to interpret. So we conducts the factor analysis on indices to extract underlying factors, and analyzes every factor to explain the information contained in this factor. More specifically, we observe the scree plot and total variance explained matrix to determine the number of factors $n_v$ in the first round, and obtain $n_v$ factors and corresponding eigenvalue $\lambda_1,\lambda_2,...,\lambda_{n_v}$ in the second round. Then we analyze the rotated component matrix and extract primary indices of each factor according to its factor loading.

 We describe the CVCL method here in detail and the framework of CVCL is presented in Fig\ref{model}. Given a dataset in which arbitrary element can be seen as combination of multiple views, i.e., $D=\{E^{(v)}\}_{v=1}^{n_v}$ with $n_v$ views, CVCL can partition the elements considering every view comprehensively without being affected by abnormal values. Denote the cardinality of dataset $D$ as $M$, and each view is a $M \times d_v$ matrix $[e_1^{(v)},e_2^{(v)},...,e_M^{(v)}]^T$ where $d_v$ is the dimension of view $v$. The framework of CVCL mainly has two components: autoencoder and contrastive learning module. Autoencoder,consisting of an encoder and a decoder, was employed to transform the raw data to semantic features of raw dataset in the pre-training stage, while contrastive learning module takes semantic features as input and partition them into $K$ clusters through contrasting temporary clusters corresponding to different views. Contrastive learning module firstly pushes semantic features of each view into a multi-layer perception which contains several linear layers and a softmax layer, and obtains a $M\times K$ matrix $H^{(v)}$ for each view as the probability distribution of each cluster. More concretely, denote $h_{rc}^{(v)}$ as the element in $r$-th row and $c$-th column, and we define $h_{rc}^{(v)}$ as the probability assigning element $e_r$ to $c$-th cluster in view $v$. However, $H^{(v)}$ matrix may suffer from a problem that the difference among $\{h_{rj}\},j=1,2,...,K$ is not significant, so a unified target distribution is appended to the multi-layer perception as shown in Eqs.\ref{unified target distribution} to enhance the discriminability of cluster assignments. Then we can obtain a $M\times K$ matrix $P^{(v)}$.
\begin{equation}\label{unified target distribution}
    p_{rc}^{(v)}=\frac{\big(h_{rc}^{(v)}\big)^2/ {\Sigma}_{i=1}^{M}h_{ic}^{(v)}}{\Sigma_{j=1}^{K}\Big(\big(h_{rj}^{(v)}\big)^2/\Sigma_{i=1}^M h_{ij}^{(v)}\Big)}
\end{equation}

For expressing the similarity between cluster assignments more precisely, we utilize Cosine Similarity as similarity calculation formula instead of the one used in CVCL(\cite{Chen_2023_ICCV}). Assume that $p_c^{(v)}$ is the $c$-th column of $P^{(v)}$. The similarity can be measured according to Eqs.\ref{cos}.
\begin{equation}\label{cos}
    s(p_c^{(v_1)},p_c^{(v_2)})=\frac{p_c^{(v_1)} \cdot p_c^{(v_2)}}{\big|\big|p_c^{(v_1)}\big|\big|_2 \ \big|\big|p_c^{(v_2)}\big|\big|_2}
\end{equation}
CVCL method simply assigns the element $e_i$ to the cluster which has maximal mean probability among all views, i.e., 
\begin{equation}\label{CVCL estimate}
    \underset{c}{\arg \max} \Big( \frac{1}{n_v}\Sigma_{v=1}^{n_v}p_{ic}^{(v)} \Big)\ .
\end{equation}
Obviously, CVCL method treats each view equally, i.e., the weight of each view is the same. But some attributes actually contribute 
to plant's growth more than others in the real world, so it's impractical to set the same weights. To enhance the performance of the CVCL method, we figure out the weight $w_1,w_2,...,w_{n_v}$ of each view through factor analysis method and modify Eqs.\ref{CVCL estimate} as:
\begin{equation}
    \underset{c}{\arg \max} \Big( \frac{1}{n_v}\Sigma_{v=1}^{n_v} w_v p_{ic}^{(v)} \Big)
\end{equation}
where weight of view $v_i$ can be calculated as:
\begin{equation}\label{eqs:weight}
    w_i=\lambda_i \Big/ \underset{i=1}{\overset{n_v}{\sum}} \lambda_i 
\end{equation}
Experiment shows that the assignment of each element is more close to practice via introducing weight of views.

To train the network efficiently, we continue to define loss function the same as CVCL's. More specifically, the total loss function is composed by three parts, i.e.,
\begin{equation}\label{eqs:loss function}
    L=L_{pre}+\alpha L_c + \beta L_a
\end{equation}
where $L_{pre}$ is the pre-training loss, $L_c$ is the cross-view contrastive loss, $L_a$ is the regularization term, and $\alpha$ and $\beta$ are parameters decided by the user. Define $f_e^{(v)}(\cdot)$ and $f_d^{(v)}(\cdot)$ as the encoder and decoder respectively. The pre-training loss expresses the reconstruction capacity of decoder and can be measured as
\begin{equation}\label{pretraining loss}
    L_{pre}=\Sigma_{v=1}^{n_v} \Sigma_{i=1}^{M} \Big|\Big| e_i^{(v)}-f_d^{(v)}\Big(f_e^{(v)}\big(e_i^{(v)}\big)\Big) \Big| \Big|_2^2
\end{equation}
Introduced to maximize the difference of single view's assignment and minimize the difference of inter-view's assignments, cross-view contrastive loss is measured as follows:
\begin{equation}\label{contrastive loss}
\begin{aligned}
    & l^{(v_1,v_2)}=\frac{1}{K} \underset{k=1}{\overset{K}{\sum}} \log \frac{ \underset{j=1,j \neq k}{\overset{K}{\sum}} e^{s\big(p_j^{(v_1)},p_k^{(v_1)}\big)} + \underset{j=1}{\overset{K}{\sum}} e ^{s\big(p_j^{(v_1)},p_k^{(v_2)}\big)} } {e^{s\big(p_k^{(v_1)},p_k^{(v_2)}\big)}},\\
    & L_c=\frac{1}{2} \underset{v_1=1}{\overset{n_v}{\sum}}\  \underset{\substack{v_2=1,\\ v_2\neq v_1} }{\overset{n_v}{\sum}}\ \ l^{(v_1,v_2)}
\end{aligned}
\end{equation}
It is straightforward that with the purpose of minimize the cross-view contrastive loss, network ought to maximize $e^{s\big(p_k^{(v_1)},p_k^{(v_2)}\big)}$ corresponding to the difference of single view's assignment and minimize $\underset{j=1,j \neq k}{\overset{K}{\sum}} e^{s\big(p_j^{(v_1)},p_k^{(v_1)}\big)} + \underset{j=1}{\overset{K}{\sum}} e ^{s\big(p_j^{(v_1)},p_k^{(v_2)}\big)}$ corresponding to the difference of inter-view's assignments meanwhile . Finally, we define $L_a$ as follows:
\begin{equation}
    \begin{aligned}
        & q_j^{(v)}=\frac{\underset{i=1}{\overset{N}{\sum} } p_{ij}^{(v)} }{N}\\
        & L_a=\underset{v=1}{\overset{n_v}{\sum}} \underset{j=1}{\overset{K}{\sum}} q_j^{(v)}\log q_j^{(v)}
    \end{aligned}
\end{equation}

Through training the network, we can obtain more precise and reasonable assignment of clusters. The rest of process is the same as mentioned in section \ref{framework}.

\section{Results}\label{results}
\begin{table*}[hbp]
\caption{Descriptive Statistics of Raw Dataset}\label{tab1}
\centering
\begin{tabular*}{\linewidth}{@{}LLLLL@{}}
\toprule
Indices & Minimum & Maximum & Mean & Standard Deviation \\ 
\midrule
Seedling Height & 14.7 & 63.2 & 37.799 & 10.0096 \\
Ground Diameter & 0.275 & 0.559 & 0.42830 & 0.082075 \\
Number of Lateral branches & 0 & 7 & 0.98 & 1.288 \\
Root Length & 4.650 & 45.900 & 18.64916 & 8.740880\\
Fresh Weight & 6.920	& 96.160 & 37.19924 & 18.483141 \\
Leaf Chlorophyll Content & 29.5 & 56.7 & 42.731 & 7.4509\\ 
\bottomrule
\end{tabular*}
\end{table*}

In this section, we represent and analyze intermediate results of each step in section \ref{methods}. And as a result, we obtain the seedling quality classification standard for Chrysanthemum, whose performance and reasonability had been examined carefully. The total experiment was conducted on a laptop computer with Intel(R) Core(R) i5-12th CPU and 32 GB RAM.

\subsection{Analysis of dataset}
At the preprocessing stage, we firstly describe basic statistical properties of raw dataset, i.e., minimum, maximum, mean, standard deviation and correlation. IBM(R) SPSS(R) is employed to do these calculation due to its excellent statistical analysis ability. As shown in Table \ref{tab1}, it is obvious that seedling height, ground diameter and leaf chlorophyll content have moderate distribution ranges, which indicates that these indices have less dispersion degree, while distribution ranges of root length and fresh weight is slightly large. Especially, the dispersion degree of number of lateral branches is extremely high, so we guess that this index is independent of other indices. 

\begin{table}[hbp]
\caption{Testing Result of Jarque-Bera Testing among indices.}\label{tab:JB}
\centering
\begin{tabular*}{\linewidth}{@{}l@{\hspace{1.5cm}}l@{}}
\hline
Index & $p-$value of\\ 
 & the hypothesis testing\\
\hline
Seedling Height & 0.1736 \\
Ground Diameter & 0.0206 \\
Number of Lateral branches & 0.0010 \\
Root Length & 0.0104\\
Fresh Weight & 0.0149 \\
Leaf Chlorophyll Content & 0.0242\\ 
\hline
\end{tabular*}
\end{table}

\begin{figure*}[tbp]
    \centering
    \includegraphics[width=0.9\textwidth]{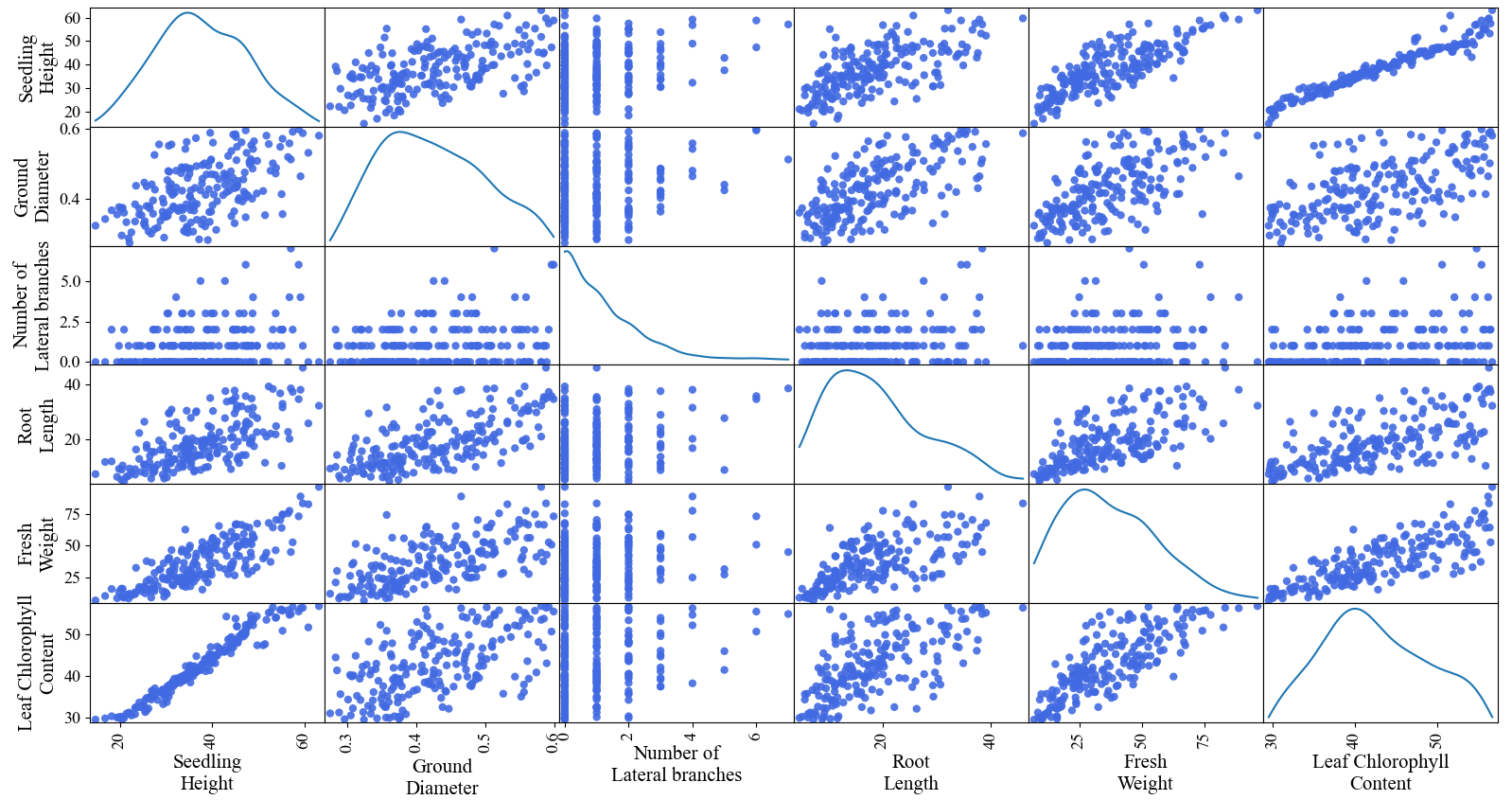}
    \caption{Scatterplot matrix of indices with kernel density estimation.}
    \label{fig:scatter}
\end{figure*}

\begin{figure*}[b]
    \centering
    \includegraphics[width=0.9\textwidth]{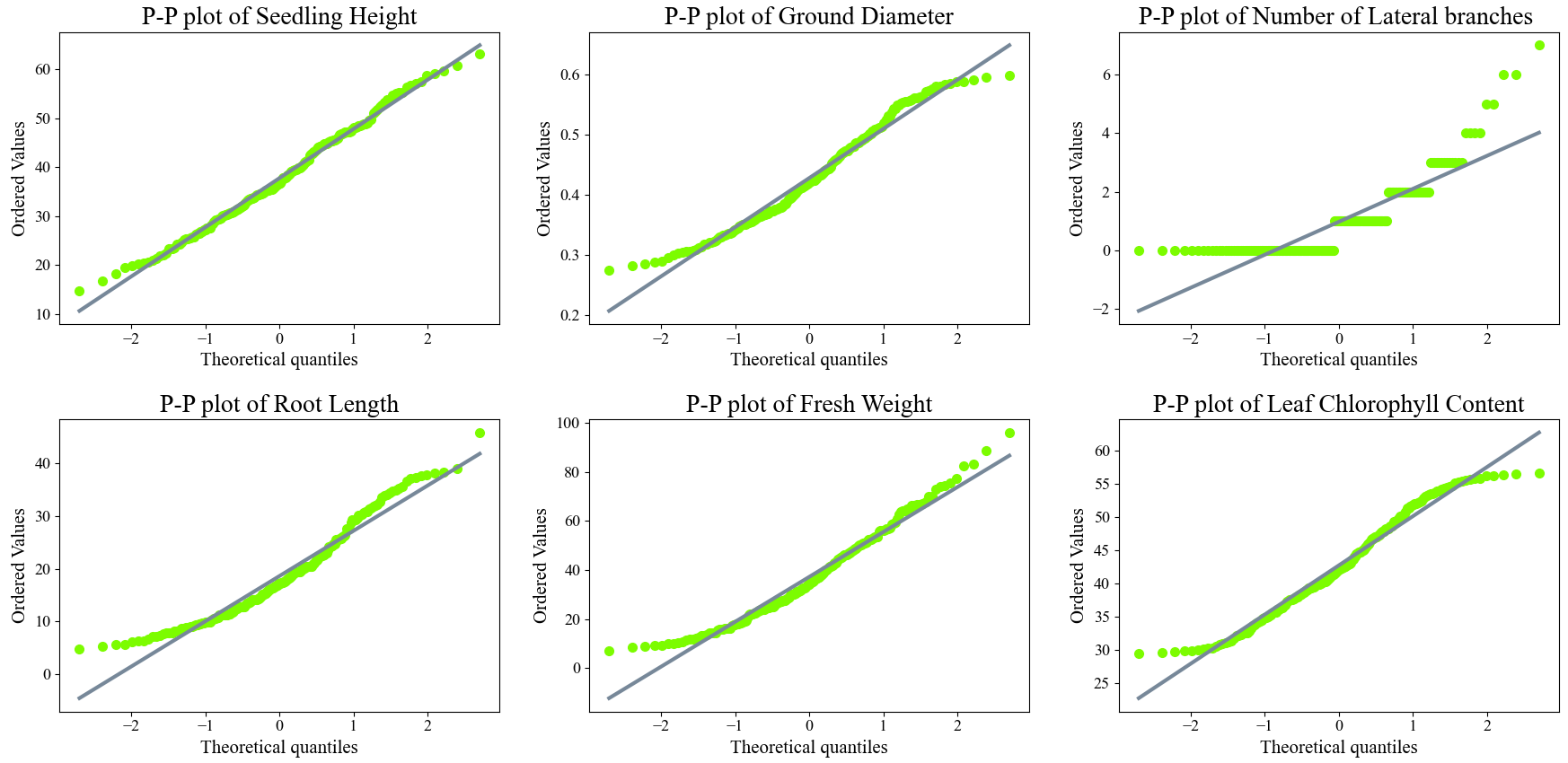}
    \caption{P-P plot of indices. If the points roughly follow a straight line, it suggests that the variable obey a normal distribution. However, if there are significant deviations from the linearity, it may indicate non-normality.}
    \label{fig:PP}
\end{figure*}

For intuition, we plot the scatterplot matrix as Fig.\ref{fig:scatter} shown. Obviously, except for number of lateral branches, each index basically obey linear correlation with other indices. Meanwhile, result of  kernel density estimation(KDE) states that indices may obey normal distribution, of course except for number of lateral branches. So we firstly tried to figure out the PCC among seedling height, ground diameter, root length, fresh weight and leaf chlorophyll content, then figure out SCC between number of lateral branches and the other 5 indices. Table.\ref{tab:JB} shows the conclusion of Jarque-Bera testing. 

\begin{figure}[h]
    \centering
    \includegraphics[width=0.8\linewidth]{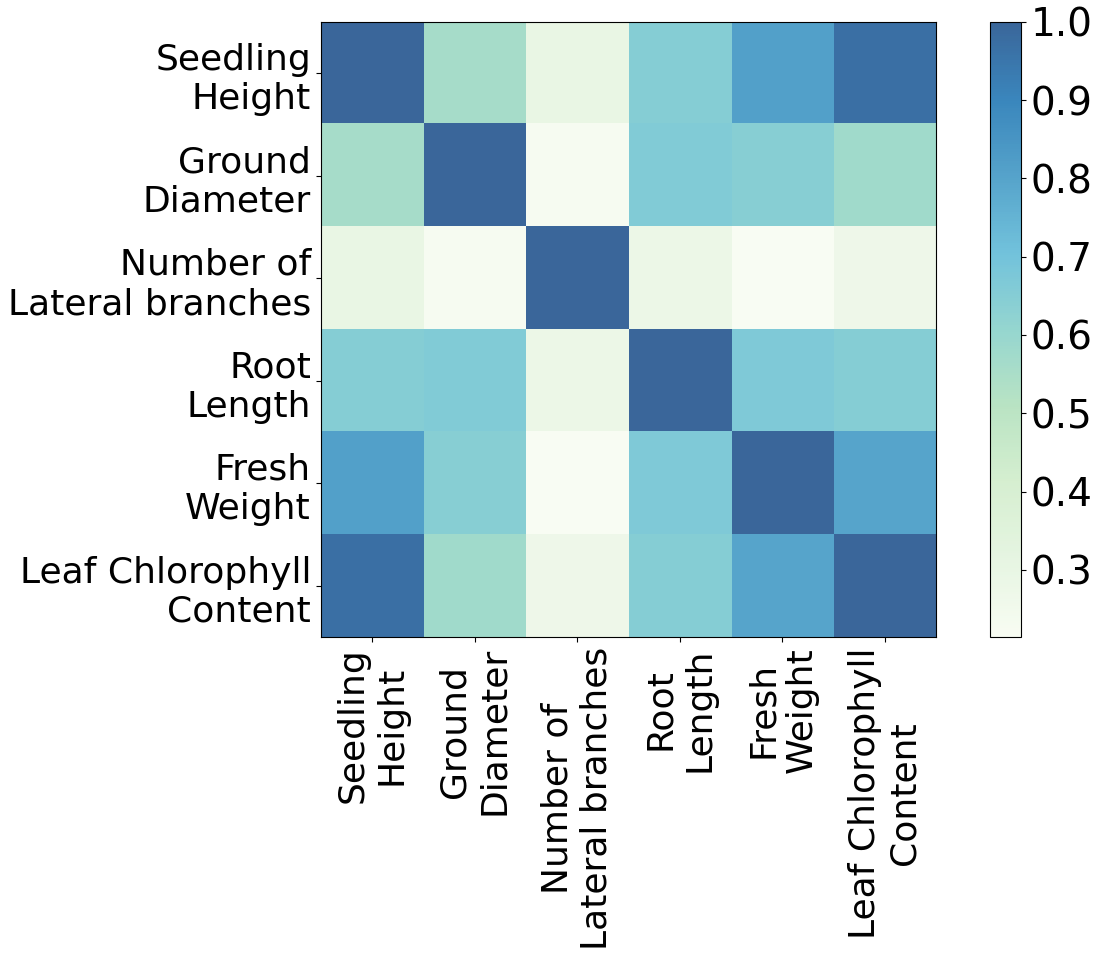}
    \caption{Correlation heat map of  indices.}
    \label{fig:heatmap}
\end{figure}

Just as we suspected, seedling height, ground diameter, root length, fresh weight and leaf chlorophyll content obey the normal distribution at 99\% confidence level since that their $p-$values are larger than $0.01$ , while number of lateral branches does not. Of course, we also plotted the Probability plot(P-P plot) in Fig.\ref{fig:PP} to estimate the distribution type roughly. The PCC and SCC of indices are presented in Table.\ref{tab:PCC} and Table.\ref{tab:SCC} respectively, and correlation heat map(Fig.\ref{fig:heatmap}) is employed to enhance visibility. 

\begin{table*}[hbp]
\caption{Pearson correlation coefficient of indices with $p-$value of the hypothesis testing. Here we define $*$ as correlation is significant at 90\% confidence level, $**$ as correlation is significant at 95\% confidence level, and $***$ as correlation is significant at 99\% confidence level.}\label{tab:PCC}
\centering
\begin{tabular*}{\linewidth}{@{}LLLLLL@{}}
\toprule
 & Seedling Height & Ground Diameter & Root Length & Fresh Weight & Leaf Chlorophyll Content \\ 
\midrule
Seedling Height & 1.0000 & 0.5579*** & 0.6488*** & 0.8111*** & 0.9716*** \\
($p-$value) & 1.0000 & 0.0000& 0.0000 & 0.0000 & 0.0000\\
Ground Diameter & 0.5579*** & 1.0000 & 0.6613*** & 0.6463*** & 0.5762***\\
($p-$value) & 0.0000 & 1.0000& 0.0000 & 0.0000 & 0.0000\\
Root Length & 0.6488*** &   0.6613***  &  1.0000  &  0.6677***  &  0.6470***\\
($p-$value) & 0.0000 & 0.0000& 1.0000 & 0.0000 & 0.0000\\
Fresh Weight & 0.8111***  &  0.6463***  &  0.6677***  &  1.0000  &  0.7983***\\
($p-$value) & 0.0000 & 0.0000& 0.0000 & 1.0000 & 0.0000\\
Leaf Chlorophyll Content & 0.9716***  &  0.5762***  &  0.6470***  &  0.7983***  &  1.0000\\
($p-$value) & 0.0000 & 0.0000& 0.0000 & 0.0000 & 1.0000\\
\bottomrule
\end{tabular*}
\end{table*}

\begin{table*}[hbp]
\caption{Spearman correlation coefficient of indices with $p-$value of the hypothesis testing. The notions are the same as described in Table.\ref{tab:PCC}}\label{tab:SCC}
\centering
\begin{tabular*}{\linewidth}{@{}LLLLLL@{}}
\toprule
 & Seedling Height & Ground Diameter & Root Length & Fresh Weight & Leaf Chlorophyll Content \\
Number of Lateral branches & 0.2956*** & 0.2234*** & 0.2758*** & 0.2138*** & 0.2710***\\
($p-$value) & 0.0000 & 0.0014 & 0.0000 & 0.0024 & 0.0001 \\
\bottomrule
\end{tabular*}
\end{table*}

Seedling height has an extremely strong correlation with leaf chlorophyll content, and a fairly strong correlation with fresh weight and root length. It conforms to common sense since that the taller seedling is, the easier it is to receive illumination, so more chlorophyll can be synthesized. And the height of the seedling affects the volume of its aboveground part, which indirectly affects the fresh weight of the seedling. In general, the growth of the aboveground part of the seedling requires a well-developed root system, so the root length is positively correlated with the height of the seedling. Ground diameter has a fairly strong correlation with root length and fresh weight. Strong roots usually have longer root system, which can obtain enough nutrients for the seedlings and also contributes to fresh weight. Especially, number of lateral branches has a very weak correlation with the other 5 indices. In practice, lateral branches will consume nutrients of seedlings, block the sunlight which affects photosynthesis, and increase the risk of diseases and pests. So it is advised to trim lateral branches regularly.

\subsection{Performance of SQCSEF using different clustering method}
It is convenient to conduct K-Means clustering with help of SPSS, so we concentrate on the CVCL method. CVCL method requires the data be divided into several views, so factor analysis is called before SQCSEF. We firstly test if indices are suitable for factor analysis and decide the number of factors. The results of KMO test and Bartlett's test is reported in Table.\ref{tab:KMO}. 
\begin{table}[h]
\caption{Result of KMO test and Bartlett's test.}\label{tab:KMO}
\centering
\begin{tabularx}{\linewidth}{@{}l@{\hspace{3.6cm}}l@{}}
\hline
KMO Measure of & Significance of   \\ 
Sampling Adequacy & Bartlett's test\\
\hline
0.807 & 0.000 \\
\hline
\end{tabularx}
\end{table}

\begin{table}[htb]
\caption{Total variances explained of factor analysis.}\label{tab:variance explained}
\centering
\begin{tabular*}{\linewidth}{@{}l@{\hspace{0.2cm}}l@{\hspace{0.2cm}}ll@{}}
\hline
Factor & Eigenvalue & Variance Explained & Cumulative\\
 & & & Variance Explained \\
\hline
1 & 3.922 & 65.372\% & 65.372\%\\
2 & 0.894 & 14.907\% & 80.279\%\\
3 & 0.607 & 10.123\% & 90.402\%\\
4 & 0.330 & 5.498\% & 95.900\%\\
5 & 0.219 & 3.650\% & 99.550\%\\
6 & 0.027 & 0.450\% & 100.000\%\\
\hline
\end{tabular*}
\end{table}

\begin{table}[htb]
\caption{Rotated component matrix of factor analysis.}\label{tab:rotated component matrix}
\centering
\begin{tabular*}{\linewidth}{@{}llll@{}}
\hline
Index & factor1 & factor2 & factor3\\
\hline
Seedling Height & 0.928 & 0.293 & 0.152\\
Ground Diameter & 0.286 & 0.892 & 0.079\\
Number of Lateral branches & 0.126 & 0.114 & 0.984\\
Root Length & 0.441 &0.752 & 0.155\\
Fresh Weight & 0.765 & 0.494 & 0.045\\
Leaf Chlorophyll Content & 0.920 & 0.309 & 0.126\\
Eigenvalue after Rotating & 2.585 & 1.799 & 1.040\\
\hline
\end{tabular*}
\end{table}

\begin{figure}[htb]
    \centering
    \includegraphics[width=0.9\linewidth]{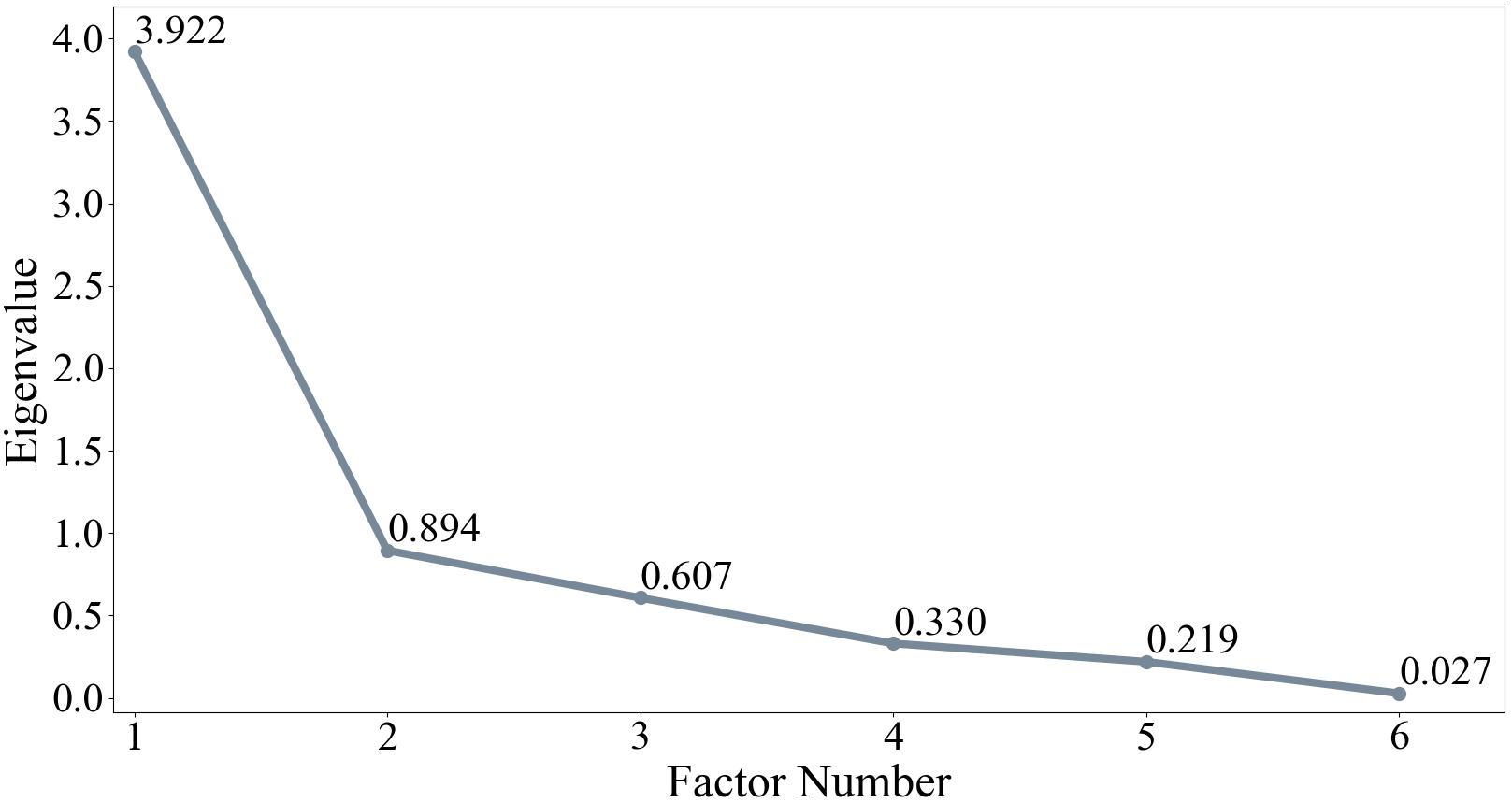}
    \caption{Scree plot of factor analysis. The Y-axis represents eigenvalue of each factor.}
    \label{fig:scree plot}
\end{figure}

Since the KMO Measure of sampling adequacy is $0.807\geq 0.80$ and significance of Bartlett's test is $0.000\leq 0.050$, factor analysis can be conducted on the raw dataset. The scree plot is plotted in Fig.\ref{fig:scree plot}, from which we observe that the curve begins to leave off after the second point, hence we consider to set the number of factors as $2$ or $3$. The total variance explained table is reported in Table.\ref{tab:variance explained}. Cumulative variance explained of the first three factors reaches $90.402\%\geq 80.000\%$, which means that these three factors can describe the raw dataset more comprehensively. To simplify the factor structure and make the relationships between factors more clearer and easier to understand, 
we rotate the factor matrix and report it in Table.\ref{tab:rotated component matrix}. In factor1, the loadings of seedling height, fresh weight and leaf chlorophyll content are relatively larger than other indices, so we suppose that factor1 shows the condition of seedling's aboveground part. In factor2, the loadings of ground diameter, root length and fresh height are relatively larger than other indices, so we suppose that factor2 shows the condition of seedling's underground part. In factor3, the loading of number of lateral branches is extremely larger than other indices, while lateral branches actually play a negative role in seedling's growth. So we suppose that factor3 shows the negative conditions for seedling's growth. Based on the result of factor analysis, We define three views of huangqiu seedling for CVCL method as (1) aboveground part view: \{seedling height, fresh height, leaf chlorophyll content\}, (2) underground part view: \{ground diameter, root length, fresh weight\} and (3) negative condition view: \{number of lateral branches\}. Weight of each view is figured out as 0.4366, 0.3317, 0.1917 respectively according to Eqs.\ref{eqs:weight}.

To train a more accurate model, we set the architecture of model as:
\begin{itemize}
    \item Each encoder is composed of an input layer which receives the data of each view, five hidden layers with size of 64, 128, 256, 512, 1024, and an output layer with size of 2048. The network of decoder is mirrored to that of corresponding encoder.
    
    \item The shared multi-layer perception contains three linear layers with size of 2048,1024,1024, and a softmax layer.
\end{itemize}

\begin{figure*}[htbp]
    \centering
    \begin{subfigure}{0.47\textwidth}
        \centering
        \includegraphics[width=1\linewidth]{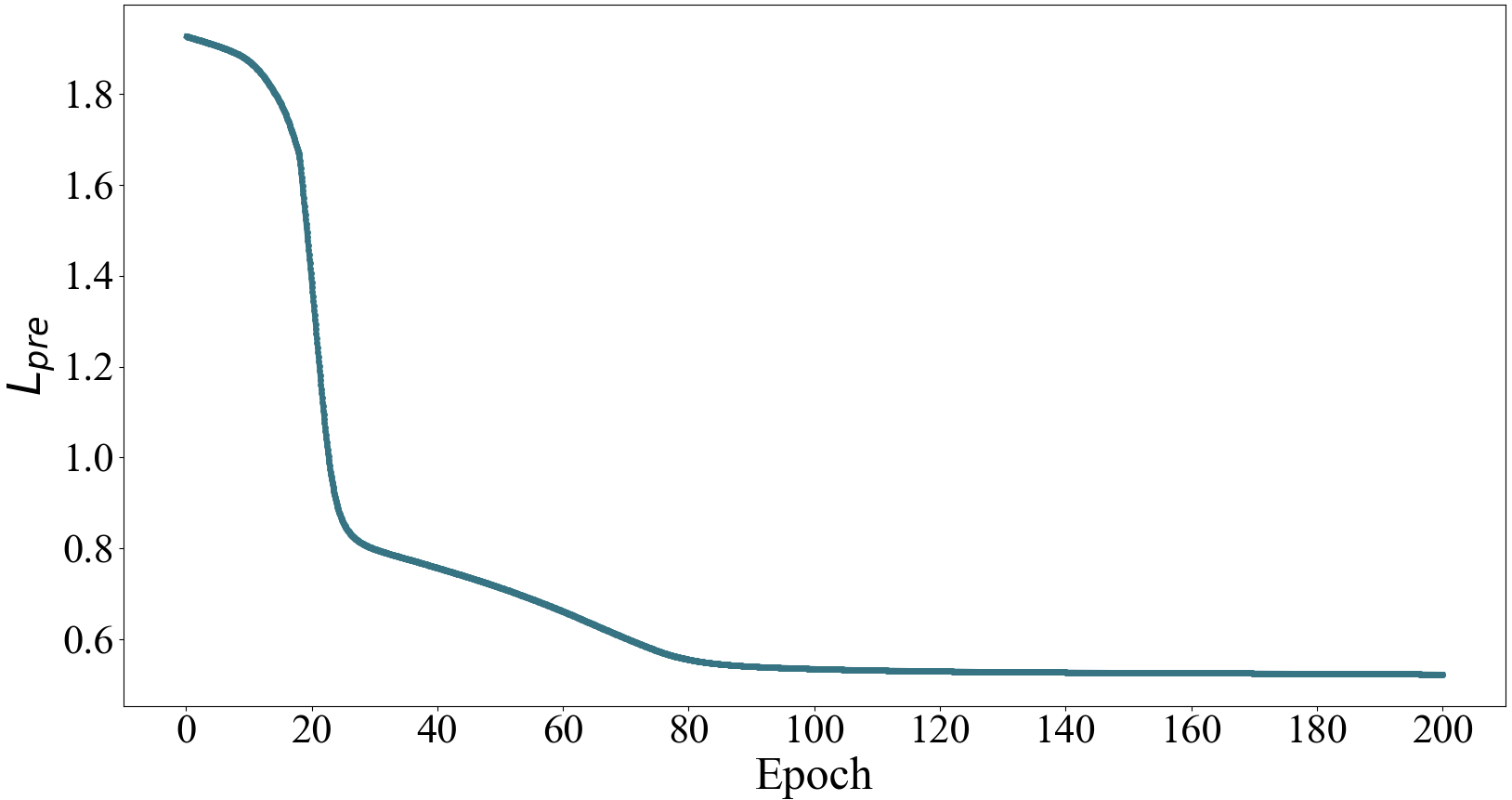}
        \caption{Loss function of Autoencoder module. The $L_{pre}$ is defined in Eqs.\ref{pretraining loss}.}
        \label{fig:loss_pre}
    \end{subfigure}
    \hfill
    \begin{subfigure}{0.47\textwidth}
        \centering
        \includegraphics[width=1\linewidth]{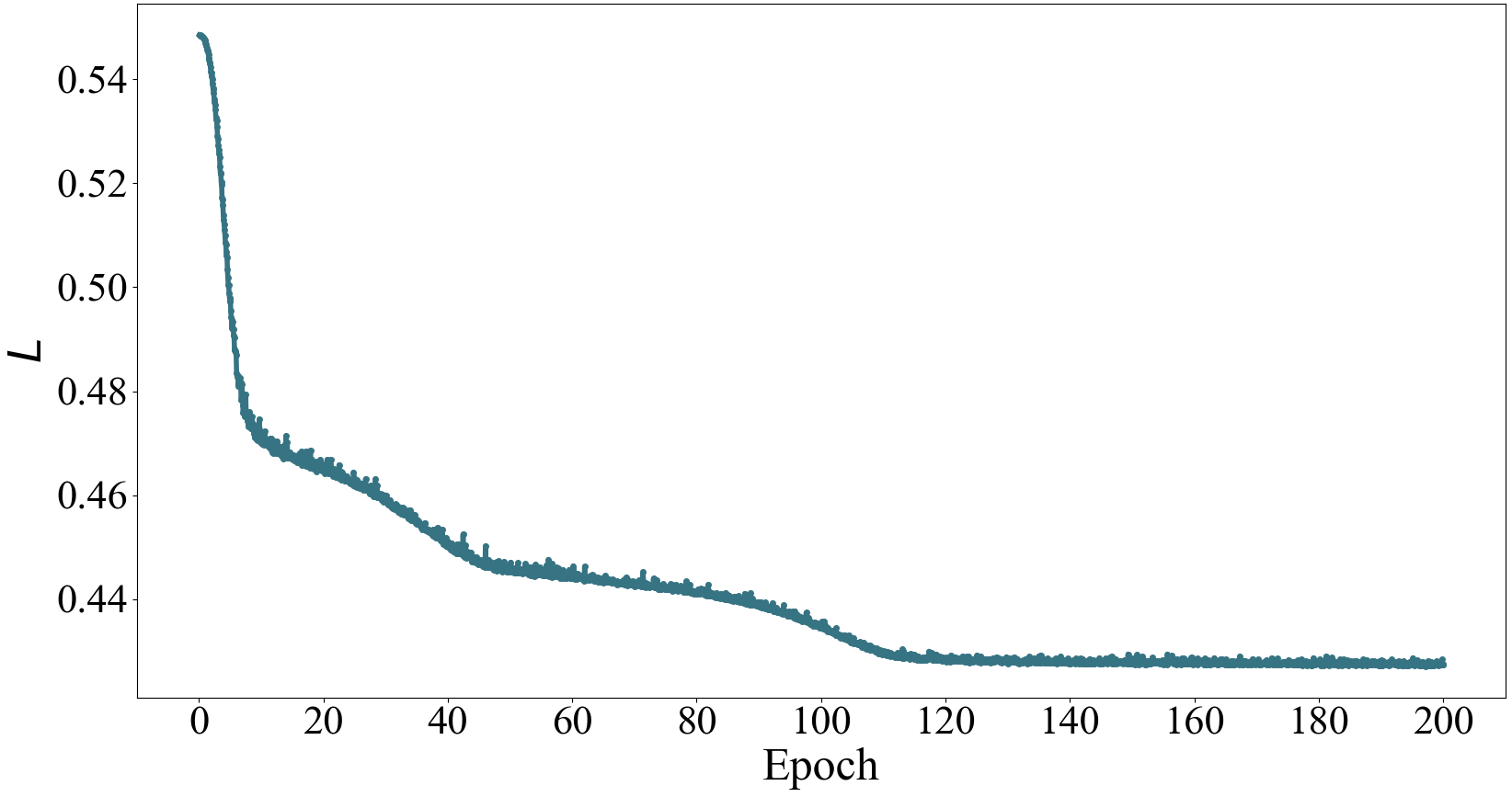}
        \caption{Loss function of Autoencoder module. The $L$ is defined in Eqs.\ref{eqs:loss function}.}
        \label{fig:loss_contra}
    \end{subfigure}
    
    \caption{Training process of CVCL.}
    \label{fig:loss}
\end{figure*}

The learning rate is set as 0.0005 and we train this model for 200 epochs. In the training stage, we firstly train the autoencoder module to minimize pre-training loss described in Eqs.\ref{pretraining loss}, then train the contrastive learning module to minimize total loss described in Eqs.\ref{eqs:loss function} . The loss function values of the model as the epoch increases is plotted in Fig.\ref{fig:loss}, which indicates that the model converges successfully. Especially, the model converges rapidly in the first 20 epochs.

We obtain three clusters for each clustering method and figure out their center and radius as shown in Table.\ref{tab:clusters}. Apparently, clusters of K-Means are more tight than that of CVCL. Then we calculate the boundary point according to Eqs.\ref{lower bound}. In the case of K-Means, the two boundary points are $(0.6620,0.6791,0.6895,0.5351,0.5426,0.7498)$ and $(0.4521,0.4620,0.7690,0.3132,0.3233,0.4745)$, while in the case of CVCL, the two boundary points are $\left(0.5875,0.5674,0.7045,0.4367,0.5067,0.6540\right)$ and $(0.4189,0.4244,0.7499,0.3064,0.2724,0.4315)$. Based on the boundary points and Eqs.\ref{normalization}, we establish seedling quality classification standard finally. We report classification standard utilizing K-Means and CVCL in Table.\ref{tab:standard kmeans} and Table.\ref{tab:standard cvcl} respectively. In order to distinguish these two classification standards, we name them as S$_{kmeans}$ and S$_{cvcl}$ respectively.

\begin{table*}[htbp]
\caption{Centers and radiuses of clusters obtained by K-Means clustering and CVCL method.}
\label{tab:clusters}
\begin{tabular*}{\tblwidth}{@{}LCLLCL@{}}
\hline
 & K-Means & & & CVCL &\\
 \cmidrule{1-3} \cmidrule{4-6} \\
 Cluster & Center  & Radius & Cluster & Center  & Radius \\
  & of Cluster & of Cluster & & of Cluster & of Cluster\\
\hline
 $C_1$ & (0.7376,0.7567,0.7683,0.5962,0.6046,0.8355) & 0.1812 & $C_1$ & (0.6727,0.6496,0.8066,0.5000,0.5801,0.7489) & 0.2070\\
 $C_2$ & (0.5039,0.5149,0.8571,0.3491,0.3603,0.5289) & 0.1374 & $C_2$ & (0.4799,0.4862,0.8592,0.3510,0.3121,0.4944) & 0.1643\\ 
 $C_3$ & (0.2842,0.2514,0.9198,0.1701,0.1521,0.2236) & 0.0828 & $C_3$ & (0.2727,0.2794,0.9152,0.1634,0.1249,0.2111) & 0.0934\\
\hline
\end{tabular*}
\end{table*}

\begin{table}[h]
\caption{S$_{kmeans}$: seedling quality classification standard utilizing K-Means clustering.}
\label{tab:standard kmeans}
\begin{tabular*}{\tblwidth}{@{}LL@{}}
\toprule
 Level & \quad Standard \\ 
\midrule
 I & Meet at least 3 of the following conditions:\\
 & Seedling height $\geq$ 46.8 cm, Ground diameter $\geq$ 0.495 cm\\
   & Number of lateral branches $\leq$ 3\\
   & Root length $\geq$ 26.7 cm, Fresh weight $geq$ 55.35 g\\
   & Leaf chlorophyll content $\geq$ 49.9 SPAD\\
 & \\
 II & Not meet the condition of level I and meet at least 3 of\\
 & the following conditions:\\
   & Seedling height $\geq$ 36.6 cm, Ground diameter $\geq$ 0.425 cm\\
   & Number of lateral branches $\leq$ 2\\
   & Root length $\geq$ 17.6 cm, Fresh weight $geq$ 35.77 g\\
   & Leaf chlorophyll content $\geq$ 42.4 SPAD\\
& \\

 III & Not meet the condition of level I,II\\
\bottomrule
\end{tabular*}
\end{table}

\begin{table}[h]
\caption{S$_{cvcl}$: Seedling quality classification standard utilizing CVCL method.}
\label{tab:standard cvcl}
\begin{tabular*}{\tblwidth}{@{}LL@{}}
\toprule
 Level & \quad Standard \\ 
\midrule
 I & Meet at least 3 of the following conditions:\\
 & Seedling height $\geq$ 43.2 cm, Ground diameter $\geq$ 0.459 cm\\
   & Number of lateral branches $\leq$ 2\\
   & Root length $\geq$ 22.7 cm, Fresh weight $geq$ 52.13 g\\
   & Leaf chlorophyll content $\geq$ 47.3 SPAD\\
 & \\
 II & Not meet the condition of level I and meet at least 3 of\\
 & the following conditions:\\
   & Seedling height $\geq$ 35.0 cm, Ground diameter $\geq$ 0.412 cm\\
   & Number of lateral branches $\leq$ 1\\
   & Root length $\geq$ 17.3 cm, Fresh weight $geq$ 31.23 g\\
   & Leaf chlorophyll content $\geq$ 41.2 SPAD\\
& \\

 III & Not meet the condition of level I,II\\
\bottomrule
\end{tabular*}
\end{table}

Traditional metric for evaluating clustering results can be divided into two categories, internal evaluation and external evaluation. Internal evaluation metrics assesses the quality of clustering results based on the internal properties of clusters, while external evaluation metrics compares clustering results with some known information such as ground truth(i.e., real assignments of samples) or expert knowledge. Evidently, we ought to employ some internal evaluation metrics, such as Silhouette Index, Calinski-Harabasz Index and Davies-Bouldin Index, to analyze the clustering effects initially. We report the result of internal evaluation in Table.\ref{tab:internal evaluation}. Due to the Silhouette index and Calinski-Harabasz index of K-means clustering are larger than that of CVCL method, while Davies-Bouldin index is smaller than CVCL's, the clusters obtained by K-Means have better tightness and higher separation degree. 

\begin{table}[h]
\caption{Internal evaluation of clusters obtained by K-Means clustering and CVCL method.}
\label{tab:internal evaluation}
\begin{tabular*}{\tblwidth}{@{}LLLL@{}}
\hline
 Clustering& Silhouette & Calinski-Harabasz & Davies-Bouldin \\ 
 Method & index & index & index\\
\hline
K-Means & 0.4489 & 138.8229 & 1.2455 \\ 
CVCL & 0.2639 & 95.6426 & 1.5846\\
\hline
\end{tabular*}
\end{table}

It seems that K-Means clustering have better performance than CVCL method. However, compared two classifications, S$_{kmeans}$ and S$_{cvcl}$, we find that S$_{cvcl}$ is more in line with reality. Each level in S$_{cvcl}$ has a wider range, which reveals that S$_{cvcl}$ can assess the sample from different views of seedling. Consider the 12th sample whose seedling height is 37.9 cm, ground diameter is 0.497 cm, number of lateral branches is 0, root length is 14.2 cm, fresh weight is 52.82 g and leaf chlorophyll content is 42.4 SPAD. S$_{kmeans}$ and S$_{cvcl}$ assigns it into level II and level I respectively. Apparently, the ground diameter and fresh weight of this sample are fairly higher, which indicates that the underground part of this seedling developed well and it has accumulated sufficient nutrients, while the seedling height, leaf content, and root length are around the average values which means that the aboveground part of the sample developed at an normal level. Meanwhile, this seedling does not have any lateral branch meaning less nutrients are wasted on useless part. So in practice this sample tends to be assigned into level I. We attribute the surprising performance of S$_{cvcl}$ to employing CVCL method as clustering method of SQCSEF. CVCL method partitions indices of seedlings into different views, and contrast assignment of each view comprehensively, which prevents the assignments from being interfered by abnormal values.


\section{Conclusions}\label{conclusion}
In this paper, we propose a simple, efficient, and general Seedling Quality Classification Standard Establishment Framework (SQCSEF) with a flexible clustering algorithm module, suitable for the classification standard establishment of most plant varieties. We then apply the SQCSEF framework to develop a classification standard for edible Chrysanthemum seedlings. To validate the generality of the SQCSEF framework, we utilize both the k-means clustering algorithm and the state-of-the-art deep clustering algorithm CVCL as the clustering algorithm module within the framework. Using a sample of 200 sets of edible Chrysanthemum 'Huangqiu' seedling data, we calculate and derive two classification standards, $S_{cvcl}$ and $S_{kmeans}$. Specifically, the CVCL method focuses on different views of the samples, effectively revealing the relationships and differences among samples from these varied perspectives. Therefore, we used factor analysis to categorize the multiple indicators of 'Huangqiu' seedlings into three views: aboveground part view, underground part view, and negative condition view. We then constructed a network comprising a 5-layer encoder and a 3-layer perceptron. After 200 epochs of training, we obtained 3 clusters, from which the final classification standard $S_{cvcl}$ was derived. Owing to CVCL's design that minimizes intra-cluster similarity and maximizes inter-cluster differences, $S_{cvcl}$ can comprehensively evaluate various sample indicators, thereby rationally classifying the samples. We are confirmed that this work will make a valuable contribution to the field of seedling classification standards.


\bibliographystyle{cas-model2-names}

\bibliography{references}

\begin{thebibliography}{26}
\expandafter\ifx\csname natexlab\endcsname\relax\def\natexlab#1{#1}\fi
\providecommand{\url}[1]{\texttt{#1}}
\providecommand{\href}[2]{#2}
\providecommand{\path}[1]{#1}
\providecommand{\DOIprefix}{doi:}
\providecommand{\ArXivprefix}{arXiv:}
\providecommand{\URLprefix}{URL: }
\providecommand{\Pubmedprefix}{pmid:}
\providecommand{\doi}[1]{\href{http://dx.doi.org/#1}{\path{#1}}}
\providecommand{\Pubmed}[1]{\href{pmid:#1}{\path{#1}}}
\providecommand{\bibinfo}[2]{#2}
\ifx\xfnm\relax \def\xfnm[#1]{\unskip,\space#1}\fi
\bibitem[{Acikgoz(2017)}]{acikgoz2017edible}
\bibinfo{author}{Acikgoz, F.}, \bibinfo{year}{2017}.
\newblock \bibinfo{title}{Edible flowers}.
\newblock \bibinfo{journal}{Journal of Experimental Agriculture International} \bibinfo{volume}{17}, \bibinfo{pages}{1--5}.
\bibitem[{Chen et~al.(2017)Chen, Lv and Zhang}]{Chen2017UnsupervisedMC}
\bibinfo{author}{Chen, D.}, \bibinfo{author}{Lv, J.}, \bibinfo{author}{Zhang, Y.}, \bibinfo{year}{2017}.
\newblock \bibinfo{title}{Unsupervised multi-manifold clustering by learning deep representation}, in: \bibinfo{booktitle}{AAAI Workshops}.
\newblock \URLprefix \url{https://api.semanticscholar.org/CorpusID:44188811}.
\bibitem[{Chen et~al.(2023)Chen, Mao, Woo and Peng}]{Chen_2023_ICCV}
\bibinfo{author}{Chen, J.}, \bibinfo{author}{Mao, H.}, \bibinfo{author}{Woo, W.L.}, \bibinfo{author}{Peng, X.}, \bibinfo{year}{2023}.
\newblock \bibinfo{title}{Deep multiview clustering by contrasting cluster assignments}, in: \bibinfo{booktitle}{Proceedings of the IEEE/CVF International Conference on Computer Vision (ICCV)}, pp. \bibinfo{pages}{16752--16761}.
\bibitem[{Clark et~al.(2000)Clark, Schalarbaum and Kormanik}]{10.1093/sjaf/24.2.93}
\bibinfo{author}{Clark, S.}, \bibinfo{author}{Schalarbaum, S.}, \bibinfo{author}{Kormanik, P.}, \bibinfo{year}{2000}.
\newblock \bibinfo{title}{{Visual Grading and Quality of 1-0 Northern Red Oak Seedlings}}.
\newblock \bibinfo{journal}{Southern Journal of Applied Forestry} \bibinfo{volume}{24}, \bibinfo{pages}{93--97}.
\newblock \URLprefix \url{https://doi.org/10.1093/sjaf/24.2.93}, \DOIprefix\doi{10.1093/sjaf/24.2.93}.
\bibitem[{Dizaji et~al.(2017)Dizaji, Herandi, Deng, Cai and Huang}]{8237874}
\bibinfo{author}{Dizaji, K.G.}, \bibinfo{author}{Herandi, A.}, \bibinfo{author}{Deng, C.}, \bibinfo{author}{Cai, W.}, \bibinfo{author}{Huang, H.}, \bibinfo{year}{2017}.
\newblock \bibinfo{title}{Deep clustering via joint convolutional autoencoder embedding and relative entropy minimization}, in: \bibinfo{booktitle}{2017 IEEE International Conference on Computer Vision (ICCV)}, pp. \bibinfo{pages}{5747--5756}.
\newblock \DOIprefix\doi{10.1109/ICCV.2017.612}.
\bibitem[{Driver(1932)}]{clustering}
\bibinfo{author}{Driver, Harold E.and~Kroeber, A.L.}, \bibinfo{year}{1932}.
\newblock \bibinfo{title}{Quantitative expression of cultural relationships}.
\newblock \bibinfo{journal}{University of California Publications in American Archaeology and Ethnology} \bibinfo{volume}{Quantitative Expression of Cultural Relationships}, \bibinfo{pages}{211--256}.
\bibitem[{Fernandes et~al.(2020)Fernandes, Casal, Pereira, Saraiva and Ramalhosa}]{doi:10.1080/87559129.2019.1639727}
\bibinfo{author}{Fernandes, L.}, \bibinfo{author}{Casal, S.}, \bibinfo{author}{Pereira, J.A.}, \bibinfo{author}{Saraiva, J.A.}, \bibinfo{author}{Ramalhosa, E.}, \bibinfo{year}{2020}.
\newblock \bibinfo{title}{An overview on the market of edible flowers}.
\newblock \bibinfo{journal}{Food Reviews International} \bibinfo{volume}{36}, \bibinfo{pages}{258--275}.
\newblock \URLprefix \url{https://doi.org/10.1080/87559129.2019.1639727}, \DOIprefix\doi{10.1080/87559129.2019.1639727}, \href{http://arxiv.org/abs/https://doi.org/10.1080/87559129.2019.1639727}{\tt arXiv:https://doi.org/10.1080/87559129.2019.1639727}.
\bibitem[{Grossnickle and MacDonald(2018)}]{grossnickle2018seedlings}
\bibinfo{author}{Grossnickle, S.C.}, \bibinfo{author}{MacDonald, J.E.}, \bibinfo{year}{2018}.
\newblock \bibinfo{title}{Why seedlings grow: influence of plant attributes}.
\newblock \bibinfo{journal}{New forests} \bibinfo{volume}{49}, \bibinfo{pages}{1--34}.
\bibitem[{Hadizadeh et~al.(2022)Hadizadeh, Samiei and Shakeri}]{hadizadeh2022chrysanthemum}
\bibinfo{author}{Hadizadeh, H.}, \bibinfo{author}{Samiei, L.}, \bibinfo{author}{Shakeri, A.}, \bibinfo{year}{2022}.
\newblock \bibinfo{title}{Chrysanthemum, an ornamental genus with considerable medicinal value: A comprehensive review}.
\newblock \bibinfo{journal}{South African Journal of Botany} \bibinfo{volume}{144}, \bibinfo{pages}{23--43}.
\bibitem[{Jiang et~al.(2015)Jiang, Le, Wan, Zhai, Hu, Xu and Xiao}]{jiang2015flower}
\bibinfo{author}{Jiang, B.}, \bibinfo{author}{Le, L.}, \bibinfo{author}{Wan, W.}, \bibinfo{author}{Zhai, W.}, \bibinfo{author}{Hu, K.}, \bibinfo{author}{Xu, L.}, \bibinfo{author}{Xiao, P.}, \bibinfo{year}{2015}.
\newblock \bibinfo{title}{The flower tea coreopsis tinctoria increases insulin sensitivity and regulates hepatic metabolism in rats fed a high-fat diet}.
\newblock \bibinfo{journal}{Endocrinology} \bibinfo{volume}{156}, \bibinfo{pages}{2006--2018}.
\bibitem[{Jingyun et~al.(2021)Jingyun, Baiyi and Baojun}]{ZHENG2021127940}
\bibinfo{author}{Jingyun, Z.}, \bibinfo{author}{Baiyi, L.}, \bibinfo{author}{Baojun, X.}, \bibinfo{year}{2021}.
\newblock \bibinfo{title}{An update on the health benefits promoted by edible flowers and involved mechanisms}.
\newblock \bibinfo{journal}{Food Chemistry} \bibinfo{volume}{340}, \bibinfo{pages}{127940}.
\newblock \URLprefix \url{https://www.sciencedirect.com/science/article/pii/S0308814620318021}, \DOIprefix\doi{https://doi.org/10.1016/j.foodchem.2020.127940}.
\bibitem[{Lloyd(1982)}]{1056489}
\bibinfo{author}{Lloyd, S.}, \bibinfo{year}{1982}.
\newblock \bibinfo{title}{Least squares quantization in pcm}.
\newblock \bibinfo{journal}{IEEE Transactions on Information Theory} \bibinfo{volume}{28}, \bibinfo{pages}{129--137}.
\newblock \DOIprefix\doi{10.1109/TIT.1982.1056489}.
\bibitem[{Lu et~al.(2016)Lu, Li and Yin}]{lu2016phytochemical}
\bibinfo{author}{Lu, B.}, \bibinfo{author}{Li, M.}, \bibinfo{author}{Yin, R.}, \bibinfo{year}{2016}.
\newblock \bibinfo{title}{Phytochemical content, health benefits, and toxicology of common edible flowers: a review (2000--2015)}.
\newblock \bibinfo{journal}{Critical Reviews in Food Science and Nutrition} \bibinfo{volume}{56}, \bibinfo{pages}{S130--S148}.
\bibitem[{Novikov et~al.(2019)Novikov, Sokolov, Drapalyuk, Zelikov and Ivetić}]{f10121064}
\bibinfo{author}{Novikov, A.}, \bibinfo{author}{Sokolov, S.}, \bibinfo{author}{Drapalyuk, M.}, \bibinfo{author}{Zelikov, V.}, \bibinfo{author}{Ivetić, V.}, \bibinfo{year}{2019}.
\newblock \bibinfo{title}{Performance of scots pine seedlings from seeds graded by colour}.
\newblock \bibinfo{journal}{Forests} \bibinfo{volume}{10}.
\newblock \URLprefix \url{https://www.mdpi.com/1999-4907/10/12/1064}, \DOIprefix\doi{10.3390/f10121064}.
\bibitem[{Pinto et~al.(2011)Pinto, Marshall, Dumroese, Davis and Cobos}]{pinto2011establishment}
\bibinfo{author}{Pinto, J.R.}, \bibinfo{author}{Marshall, J.D.}, \bibinfo{author}{Dumroese, R.K.}, \bibinfo{author}{Davis, A.S.}, \bibinfo{author}{Cobos, D.R.}, \bibinfo{year}{2011}.
\newblock \bibinfo{title}{Establishment and growth of container seedlings for reforestation: A function of stocktype and edaphic conditions}.
\newblock \bibinfo{journal}{Forest Ecology and Management} \bibinfo{volume}{261}, \bibinfo{pages}{1876--1884}.
\bibitem[{Qian(2023)}]{Qian_2023_ICCV}
\bibinfo{author}{Qian, Q.}, \bibinfo{year}{2023}.
\newblock \bibinfo{title}{Stable cluster discrimination for deep clustering}, in: \bibinfo{booktitle}{Proceedings of the IEEE/CVF International Conference on Computer Vision (ICCV)}, pp. \bibinfo{pages}{16645--16654}.
\bibitem[{Rop et~al.(2012)Rop, Mlcek and Jurikova}]{molecules17066672}
\bibinfo{author}{Rop, O.}, \bibinfo{author}{Mlcek, J.}, \bibinfo{author}{Jurikova}, \bibinfo{year}{2012}.
\newblock \bibinfo{title}{Edible flowers—a new promising source of mineral elements in human nutrition}.
\newblock \bibinfo{journal}{Molecules} \bibinfo{volume}{17}, \bibinfo{pages}{6672--6683}.
\newblock \URLprefix \url{https://www.mdpi.com/1420-3049/17/6/6672}.
\bibitem[{Rose et~al.(1990)Rose, Carlson and Morgan}]{rose1990target}
\bibinfo{author}{Rose, R.}, \bibinfo{author}{Carlson, W.C.}, \bibinfo{author}{Morgan, P.}, \bibinfo{year}{1990}.
\newblock \bibinfo{title}{The target seedling concept}, in: \bibinfo{booktitle}{Proceedings of Combined Meeting of the Western Forest Nursery Associations, Roseburg}, pp. \bibinfo{pages}{1--8}.
\bibitem[{Spaargaren and van Geest(2018)}]{Spaargaren2018}
\bibinfo{author}{Spaargaren, J.}, \bibinfo{author}{van Geest, G.}, \bibinfo{year}{2018}.
\newblock \bibinfo{title}{Chrysanthemum}. \bibinfo{publisher}{Springer International Publishing}, \bibinfo{address}{Cham}.
\newblock pp. \bibinfo{pages}{319--348}.
\newblock \URLprefix \url{https://doi.org/10.1007/978-3-319-90698-0_14}, \DOIprefix\doi{10.1007/978-3-319-90698-0_14}.
\bibitem[{Sutton(1980)}]{sutton1980evaluation}
\bibinfo{author}{Sutton, R.}, \bibinfo{year}{1980}.
\newblock \bibinfo{title}{Evaluation of planting stock quality}.
\newblock \bibinfo{journal}{NZJ For Sci} \bibinfo{volume}{10}, \bibinfo{pages}{293--300}.
\bibitem[{Tian et~al.(2018)Tian, Li, Li, Zhi, Li, Tang, Yang, Yin and Ming}]{tian2018protective}
\bibinfo{author}{Tian, Y.}, \bibinfo{author}{Li, Y.}, \bibinfo{author}{Li, F.}, \bibinfo{author}{Zhi, Q.}, \bibinfo{author}{Li, F.}, \bibinfo{author}{Tang, Y.}, \bibinfo{author}{Yang, Y.}, \bibinfo{author}{Yin, R.}, \bibinfo{author}{Ming, J.}, \bibinfo{year}{2018}.
\newblock \bibinfo{title}{Protective effects of coreopsis tinctoria flowers phenolic extract against d-galactosamine/lipopolysaccharide-induced acute liver injury by up-regulation of nrf2, ppar$\alpha$, and ppar$\gamma$}.
\newblock \bibinfo{journal}{Food and chemical toxicology} \bibinfo{volume}{121}, \bibinfo{pages}{404--412}.
\bibitem[{Wilcox(2017)}]{2017i}
\bibinfo{author}{Wilcox, R.}, \bibinfo{year}{2017}.
\newblock in: \bibinfo{booktitle}{Introduction to Robust Estimation and Hypothesis Testing (Fourth Edition)}. \bibinfo{edition}{fourth edition} ed.. \bibinfo{publisher}{Academic Press}. Statistical Modeling and Decision Science, pp. \bibinfo{pages}{i--iii}.
\newblock \URLprefix \url{https://www.sciencedirect.com/science/article/pii/B9780128047330000147}, \DOIprefix\doi{https://doi.org/10.1016/B978-0-12-804733-0.00014-7}.
\bibitem[{Xie et~al.(2016)Xie, Girshick and Farhadi}]{10.5555/3045390.3045442}
\bibinfo{author}{Xie, J.}, \bibinfo{author}{Girshick, R.}, \bibinfo{author}{Farhadi, A.}, \bibinfo{year}{2016}.
\newblock \bibinfo{title}{Unsupervised deep embedding for clustering analysis}, in: \bibinfo{booktitle}{Proceedings of the 33rd International Conference on International Conference on Machine Learning - Volume 48}, \bibinfo{publisher}{JMLR.org}. p. \bibinfo{pages}{478–487}.
\bibitem[{Yamamoto et~al.(2015)Yamamoto, Yamane, Oishi, Shimizu, Tadaishi and Kobayashi-Hattori}]{yamamoto2015chrysanthemum}
\bibinfo{author}{Yamamoto, J.}, \bibinfo{author}{Yamane, T.}, \bibinfo{author}{Oishi, Y.}, \bibinfo{author}{Shimizu, M.}, \bibinfo{author}{Tadaishi, M.}, \bibinfo{author}{Kobayashi-Hattori, K.}, \bibinfo{year}{2015}.
\newblock \bibinfo{title}{Chrysanthemum promotes adipocyte differentiation, adiponectin secretion and glucose uptake}.
\newblock \bibinfo{journal}{The American Journal of Chinese Medicine} \bibinfo{volume}{43}, \bibinfo{pages}{255--267}.
\bibitem[{Zaerr(2003)}]{zaerr1985role}
\bibinfo{author}{Zaerr, J.B.}, \bibinfo{year}{2003}.
\newblock \bibinfo{title}{The role of biochemical measurements in evaluating vigor}.
\bibitem[{Zhang et~al.(2023)Zhang, Xu, Wang, Long and Gao}]{10349963}
\bibinfo{author}{Zhang, H.}, \bibinfo{author}{Xu, H.}, \bibinfo{author}{Wang, X.}, \bibinfo{author}{Long, F.}, \bibinfo{author}{Gao, K.}, \bibinfo{year}{2023}.
\newblock \bibinfo{title}{A clustering framework for unsupervised and semi-supervised new intent discovery}.
\newblock \bibinfo{journal}{IEEE Transactions on Knowledge and Data Engineering} , \bibinfo{pages}{1--14}\DOIprefix\doi{10.1109/TKDE.2023.3340732}.

\end{thebibliography}

\end{sloppypar}
\end{document}